\documentclass[10pt,twocolumn,letterpaper]{article}
\usepackage{iccv}
\usepackage{times}
\usepackage{epsfig}
\usepackage{graphicx}
\usepackage{amsmath}
\usepackage{amssymb}

\usepackage{float}
\usepackage{lipsum}
\usepackage{stfloats}
\usepackage{multicol}
\usepackage{multirow}
\usepackage{bm}
\usepackage{etoolbox}

\usepackage{array}
\usepackage{tabulary}
\usepackage[table,dvipsnames]{xcolor} 
\usepackage{paralist}
\usepackage{booktabs}
\usepackage{adjustbox}
\usepackage[shortlabels]{enumitem}
\usepackage{tabu} 

\usepackage{caption}
\usepackage{subcaption}
\usepackage[pagebackref=true,breaklinks=true,letterpaper=true,colorlinks,bookmarks=false,citecolor=blue,linkcolor=red]{hyperref}
\usepackage{icomma} 
\usepackage{arydshln}

\definecolor{gray}{rgb}{0.3,0.3,0.3}
\definecolor{blue}{rgb}{0,0.5,1}
\definecolor{mask_red}{rgb}{1,0,0.8}
\definecolor{green}{rgb}{0.2,1,0.2}
\definecolor{rblue}{rgb}{0,0,1}
\definecolor{lightblue}{HTML}{6495ed}
\definecolor{lightred}{HTML}{F19C99}

\definecolor{graytablerow}{gray}{0.6}

\usepackage{tikz}

\iccvfinalcopy 

\ificcvfinal\fi

\begin{document}

%
\title{Open Scene Understanding: Grounded Situation Recognition Meets Segment Anything for Helping People with Visual Impairments}

\author{
Ruiping Liu$^1$,
~~Jiaming Zhang$^{1,}$\thanks{Corresponding author (e-mail: {\tt jiaming.zhang@kit.edu}).},
~~Kunyu Peng$^1$,
~~Junwei Zheng$^1$,
~~Ke Cao$^1$,
~~Yufan Chen$^1$,\\
Kailun Yang$^2$,
~~Rainer Stiefelhagen$^1$\\
\normalsize
$^1$Karlsruhe Institute of Technology,
\normalsize
~~$^2$Hunan University
}

\maketitle
\ificcvfinal\thispagestyle{empty}\fi

\begin{abstract}
Grounded Situation Recognition (GSR) is capable of recognizing and interpreting visual scenes in a contextually intuitive way, yielding salient activities (verbs) and the involved entities (roles) depicted in images. In this work, we focus on the application of GSR in assisting people with visual impairments (PVI). However, precise localization information of detected objects is often required to navigate their surroundings confidently and make informed decisions. For the first time, we propose an Open Scene Understanding (OpenSU) system that aims to generate pixel-wise dense segmentation masks of involved entities instead of bounding boxes. Specifically, we build our OpenSU system on top of GSR by additionally adopting an efficient Segment Anything Model (SAM). Furthermore, to enhance the feature extraction and interaction between the encoder-decoder structure, we construct our OpenSU system using a solid pure transformer backbone to improve the performance of GSR. In order to accelerate the convergence, we replace all the activation functions within the GSR decoders with GELU, thereby reducing the training duration. In quantitative analysis, our model achieves state-of-the-art performance on the SWiG dataset. Moreover, through field testing on  dedicated assistive technology datasets and application demonstrations, the proposed OpenSU system can be used to enhance scene understanding and facilitate the independent mobility of people with visual impairments. Our code will be available at \href{https://github.com/RuipingL/OpenSU}{OpenSU}.
\end{abstract}


\section{Introduction}
\label{sec:intro}
{Scene understanding is a fundamental computer vision technology that interprets visual scenes and is widely used in various scenarios, such as autonomous driving~\cite{liu2022transkd}, robotics~\cite{zheng2023materobot}, assistive technology~\cite{zhang2021trans4trans}, \textit{etc.}
In this work, we mainly explore scene understanding for helping People with Visual Impairments~(PVI).} 
PVI often have difficulty interpreting their surroundings correctly due to the lack of visual cues. Understanding a scene through tactile exploration proves insufficient and potentially hazardous. Often people need to recognize the whole scene at first glance, then gaze at each object, sort out their relationships, and react to the scene~\cite{kahneman2003cognitivesystem}. 
An assistive system providing scene descriptions and object locations could be very useful for visually impaired people to get a better understanding of the environment and thus lead towards more autonomy in many daily situations. 
Existing assistance systems~\cite{zhang2021trans4trans, zheng2023materobot} focus on parsing the given image into pixel-level predictions with class names.
\begin{figure}
    \centering
    \includegraphics[width=0.5\textwidth]{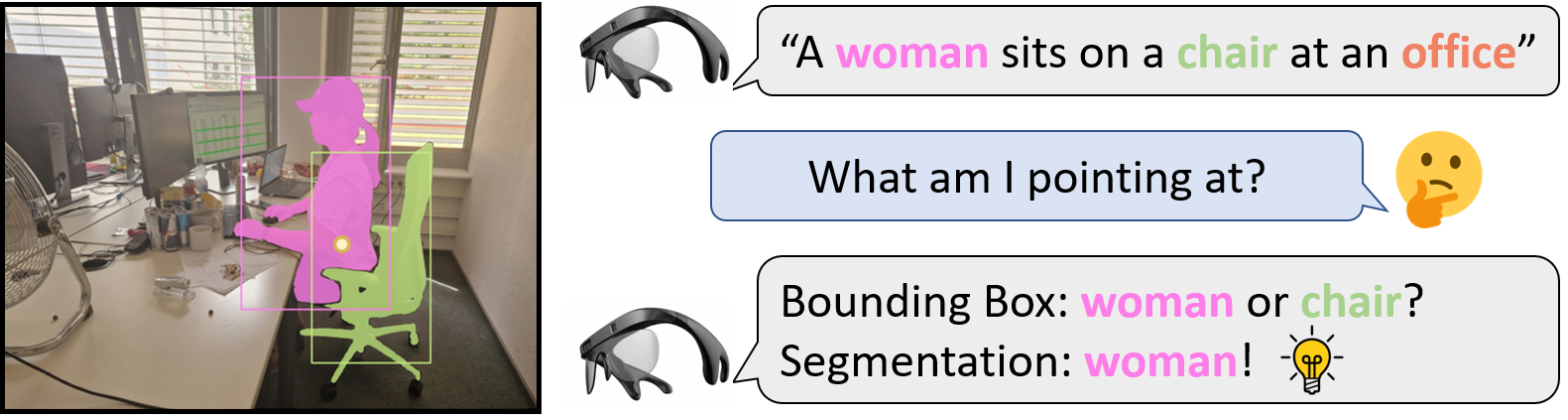}
    \vskip -2ex
    \caption{\textbf{OpenSU} generates image caption and segmentation maps of salient objects, while GSR outputs overlapping bounding boxes. The yellow point in the overlapping part of the bounding boxes represents the region of interest.}
    \label{fig:bb_vs_seg}
    \vskip -3ex
\end{figure}
However, only knowing the class name of objects -- without having any description between objects -- limits their ability to convey rich and comprehensive information about the scene. Grounded Situation Recognition (GSR)~\cite{pratt2020swig} is a verb-oriented task, which can predict the complete information of the scene, \eg, activity and entity information. 
Nevertheless, GSR models have limitations in object localization through bounding boxes, which lack fine-grained positional information and may result in ambiguous interaction, such as overlapping bounding boxes.

\begin{figure}[t!]
\subfloat[Bounding boxes overlaping\label{subfig:bbx_overlap}]{%
  \includegraphics[width=0.255\textwidth]{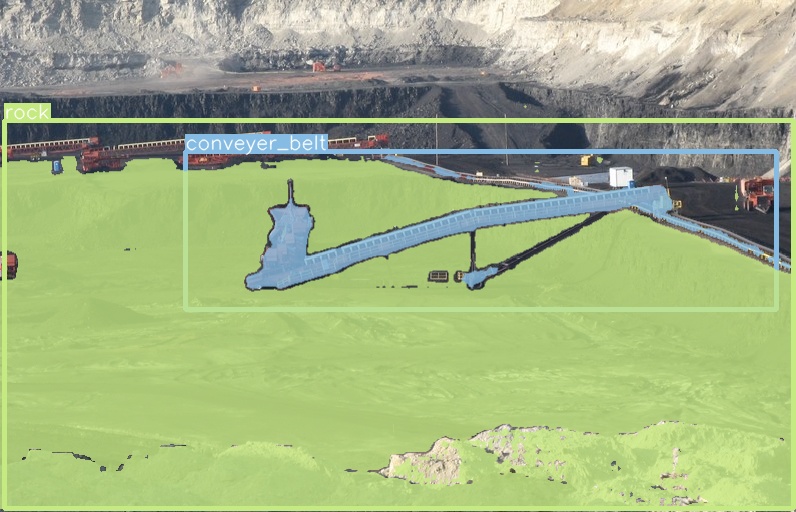}
}
\subfloat[Multiple objects in a BB \label{subfig:multiobj}]{%
  \includegraphics[width=0.218\textwidth]{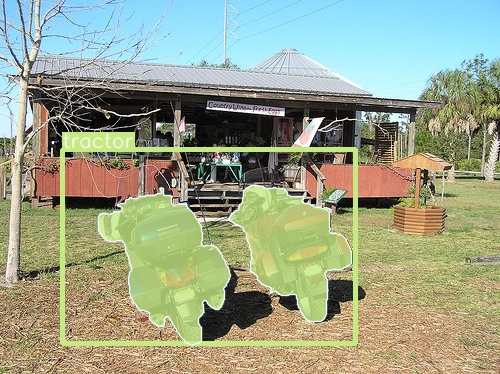}
}
\vskip -1ex
\caption{\textbf{Challenging cases} of using bounding boxes (BBs) for object localization. %
}
\label{fig:BB_Difficulties}
\vskip -2ex
\end{figure}

Recently, the advent of foundation models such as ChatGPT, GPT-3~\cite{brown2020gpt3} and Segment Anything Model (SAM)~\cite{kirillov2023segment} has made significant contributions to the advancement of Artificial General Intelligence (AGI). However, there is still a lack of exploration regarding the application of these models in downstream tasks, particularly in the realm of assistive technologies for PVI. The research question arises: \emph{how are foundation models applicable to assistance systems for helping people with visual impairments?}

Based on the aforementioned viewpoints, we introduce \textbf{Open Scene Understanding (OpenSU)} system (in Fig.~\ref{fig:bb_vs_seg}), which consists of Grounded Situation Recognition (GSR)~\cite{pratt2020swig} and Segment Anything (SAM)~\cite{kirillov2023segment}. Our OpenSU system can generate the scene description message as GSR while additionally producing disjoint segmentation masks of the objects to alleviate the confusion of object localization. 
Specifically, GSR outputs structured semantic descriptions of images, including the main activity, entities of the predefined roles (\eg, agent, tool, place), 
and the bounding-box groundings of entities. FrameNet~\cite{baker1998framenet} is the lexical-semantic resource that maps every verb to a set of semantic roles. Additionally, a preliminary frame is provided for each verb to facilitate the integration of roles, allowing for the subsequent insertion of predicted nouns into those roles. 
Besides, to achieve a wide range of open-set scene description capabilities, SWiG~\cite{pratt2020swig} dataset with over $500$ verb and $10K$ entity classes is applied to train the GSR model. 
Compared to image captioning, GSR only considers the significant words, while image captioning also considers the word order and nonessential words, like the articles. Therefore, GSR yields more efficient descriptions and does not face the evaluation challenges~\cite{anderson2016spice,vedantam2015cider} that image captioning encounters. 

\begin{figure}
    \centering
    \includegraphics[width=0.50\textwidth]{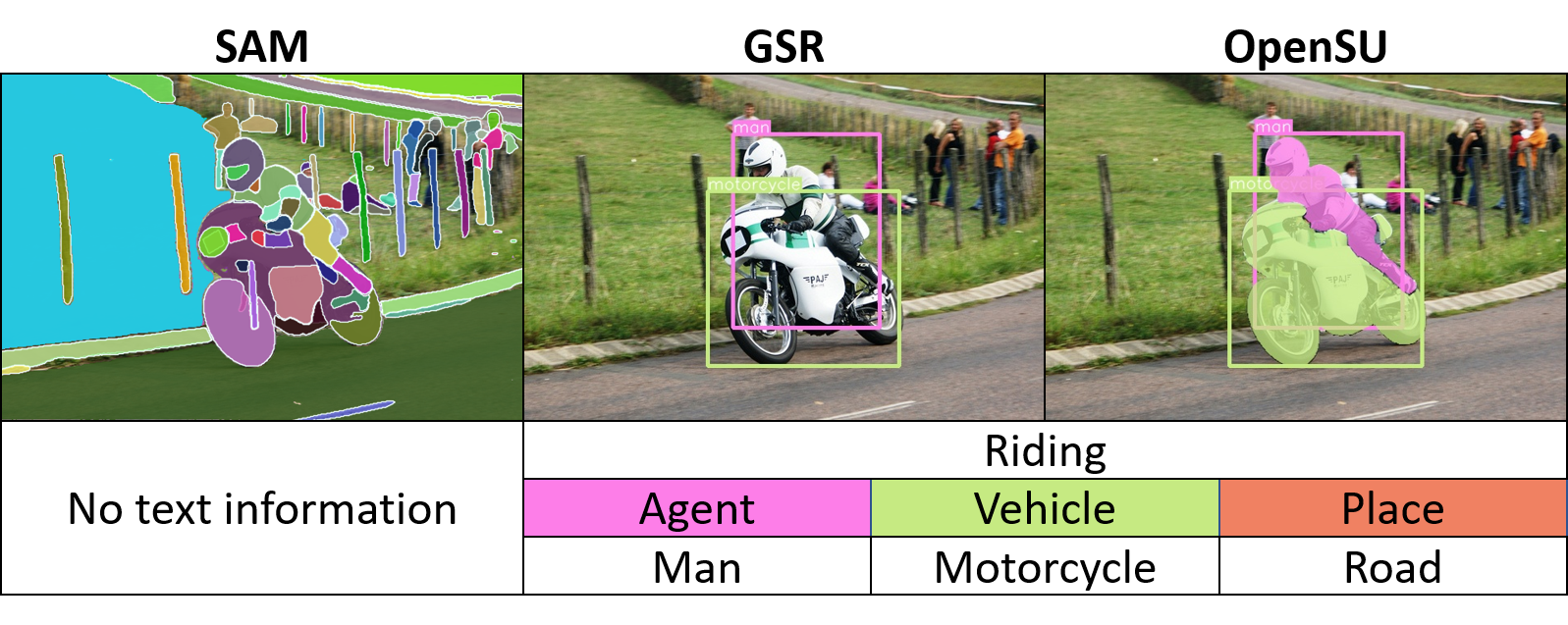}
    \vskip -1ex
    \caption{\textbf{Comparison among scene understanding methods}, including Segment Anything Model (SAM)~\cite{kirillov2023segment}, Grounded Situation Recognition (GSR)~\cite{cho2022coformer}, and our OpenSU. Our system provides the description: ``A \textit{man} rides the \textit{motorcycle} at a \textit{road}'', and semantic masks for more precise localization than overlapping bounding boxes. 
    }
    \label{fig:comparison_tasks}
    \vskip -2ex
\end{figure}

As mentioned before, ambiguous interaction occurs when people attempt to localize the objects with bounding boxes generated by GSR models, as shown in Fig.~\ref{fig:BB_Difficulties}. 
The positional relationship between objects can lead to overlapping bounding boxes (in Fig.~\ref{fig:BB_Difficulties}(a)), or a single bounding box may encompass multiple similar objects (in Fig~\ref{fig:BB_Difficulties}(b)). These challenging cases may cause confusion about object localization for PVI.
Furthermore, the coarse bounding boxes to the object outlines result in the inclusion of background elements and an incomplete representation of the objects. In the depicted situations, the system suffers from confusion to extract the semantic information from the bounding box annotated scenes. To eliminate the localization limitation, SAM takes the bounding boxes as prompt and produces the accurate segmentation masks of the instances. Fig.~\ref{fig:comparison_tasks} showcases the visual and linguistic outputs of SAM, GSR, and OpenSU. SAM generates precise masks without semantic information or text messages. GSR predicts the activities, related objects, and bounding boxes. OpenSU yields both an image description and segmentation masks, enhancing the overall scene understanding.

In our approach, first, the description based on GSR and the segmentation masks of salient objects is presented by our OpenSU system. The user is then asked to specify a region of interest to get further information. To this end, we propose various potential region indication methods.
Subsequently, the system returns the relevant semantic information corresponding to the defined region, so as to assist the user to localize the object of interest. 

About the experiment, we validate the performance of our GSR model on the SWiG dataset~\cite{pratt2020swig}, demonstrating a notable improvement of ${+}3.47\%$ over the previous state-of-the-art GSRFormer model~\cite{cheng2022gsrformer}.
The vanilla SAM suffers from large model sizes and latency. We explore different SAM variations~\cite{kirillov2023segment,zhang2023mobilesam} in order to generate segmentation maps efficiently. Compared to the counterpart Grounded-SAM\footnote{\url{https://github.com/IDEA-Research}}, a stack of foundation models, our system realizes real-time Open Scene Understanding, having $0.7s$ \textit{vs.} $74.04s$ processing time per image of the baseline.

To summarize, we present the following contributions:
\begin{compactitem}
\item We design an \emph{Open Scene Understanding (OpenSU)} system, allowing people with visual impairments to perceive the entirety of the scene and retrieve object information in a specified direction.

\item To achieve real-time OpenSU, we propose the state-of-the-art GSR model 
which combines the Swin transformer and CoFormer architecture, and uses GELU to shorten the convergence duration.

\item We for the first time, adopt and test two variations of the Segment Anything Model (SAM) in assistive technology for helping people with visual impairments.

\item The effectiveness and efficiency of the OpenSU system are verified by various experiments and field tests. 
\end{compactitem}

\section{Related Work}
\label{sec:related_work}
\subsection{Semantic Segmentation}
Semantic segmentation aims to partition a given image into several regions according to the semantic meaning, which tends to classify each pixel into its corresponding semantic class. Diverse research fields, \eg, autonomous driving~\cite{cheng2021per,fan2021rethinking, peng2022mass, strudel2021segmenter, xie2021segformer, xu2022groupvit, zheng2021rethinking}, medical image analysis~\cite{li2020shape, qiu2019semantic, xia2020synthesize}, and assistance for visually impaired people~\cite{duh2020veye, li2017situation, liu2021hida, tian2021dynamic_crosswalk, zhang2021trans4trans, zheng2023materobot, zou2023realtime_passable}, rely heavily on semantic segmentation to provide a deep understanding of the scenario to achieve better assistance.
Most of the existing semantic segmentation works~\cite{ liu2021hida, xie2021segformer,zhang2021trans4trans, zheng2021rethinking} are close-set, where the prediction space is limited by the class types provided in the training set.
With the rapid development of vision-language model~\cite{du2022learning, li2023blip}, \eg, CLIP~\cite{ilharco_gabriel_2021_5143773, Radford2021LearningTV} and BLIP~\cite{li2022blip}, and segmentation foundation models, \eg, SAM~\cite{kirillov2023segment}, open-set semantic segmentation~\cite{liang2023open, ma2022open, nunes2022conditional, uhlemeyer2022towards, xu2023learning} has played an increasingly important role in the computer vision community, which benefits from the ability to classify the pixels even for the class type which is not involved during the training and thereby relies less on the expensive pixel-level annotation on the new classes.
{Unlike existing systems, we for the first time, introduce SAM-based scene understanding for helping visually impaired people, which largely enhances the practicability of vision-based assistance in real-world unconstrained everyday scenarios.}

\subsection{Scene Understanding}
Scene understanding concentrates on comprehending the existing elements and their corresponding relationships within a given image, which results in the analysis of the status of different objects, the attributes of the agents, and their interactions.
Scene understanding enables the deep learning model to interpret the visual data, which benefits diverse practical usages~\cite{azuma2022scanqa,  liao2022unsupervised, ma2022both,wu2022p2t, zhou2022mtanet}, \eg, assistance for visually impaired~\cite{li2020cross_safe, liu2021hida, tian2021dynamic_crosswalk,  zhang2021trans4trans} and autonomous driving~\cite{chen2019towards, min2022traffic, sakaridis2021acdc}, where Grounded Situation Recognition (GSR)~\cite{cho2022coformer} is one of the most essential scene understanding tasks, which predicts the entities, the activities and the bounding boxes groundings for the corresponding entities.
SituFormer~\cite{wei2021rethinking} makes use of a coarse-to-fine verb model and a transformer-based Noun model to achieve GSR.
Cho~\etal~\cite{cho2022coformer} propose a collaborative glance-gaze transformer where the glance transformer is used to predict the main activity and the gaze transformer is designed for the estimation of entities.
Cheng~\etal~\cite{cheng2022gsrformer} propose GSRFormer to harvest structural-semantic description of a given image.
In this work, we for the first time propose a grounding situation segmentation system to achieve dense segmentation of the entities inside their bounding boxes.
We leverage Swin as the feature extraction backbone and SAM as the segmentation mask generator to achieve a superior scene understanding. The convergence is accelerated by using the GELU activation function in the whole framework. 

\subsection{Visual Assistance Systems}
Visual assistance systems are used to provide essential environmental information to achieve the navigation of the visually impaired through wearable sensors~\cite{ chen2021wearable, li2022sensing, lin2019deep_learning} and environment perceiving sensors~\cite{ duh2020veye,liu2021hida, tian2021dynamic_crosswalk, zhang2021trans4trans}.
Semantic segmentation is a well-established technique used to assist the visually impaired.
Lin~\etal~\cite{lin2019deep_learning} propose a learning-based wearable system to achieve the navigation for visually impaired.
Apart from semantic segmentation techniques, V-Eye~\cite{duh2020veye} makes use of a global localization method to pursue a better scene understanding, while the outdoor walking guide system~\cite{hsieh2020outdoor_walking_guide} leverages depth information.
Liu~\etal~\cite{liu2021hida} propose HIDA to make use of instance semantic segmentation techniques with LiDAR sensor.
The assistance for the visually impaired is likely to be involved in different traffic situations.
Tian~\etal~\cite{tian2021dynamic_crosswalk} and Li~\etal~\cite{li2020cross_safe} concentrate on handling the crosswalk situation, which involves the segmentation of objects and prediction of the traffic light status, while Zou~\etal~\cite{zou2023realtime_passable} focus on real-time passable area segmentation. Most of the above-mentioned works rely on Convolutional Neural Networks (CNN) as feature extractors.
Zhang~\etal~\cite{zhang2021trans4trans} propose Trans4Trans to leverage transformers to serve visual assistance systems.
Ma~\etal~\cite{ma2023eos} propose a robot system to achieve wayfinding tasks for the visually impaired.
The work of ``I am the follower, also the boss''~\cite{zhang2023follower} uses machine forms of a guiding robot and anatomy from different stages to achieve visually impaired assistance.
Zheng~\etal~\cite{zheng2023materobot} focus on material recognition in wearable robotics.
Apart from the existing works with user studies, there are still well-established works that only concentrate on segmentation without user study~\cite{cao2020rapid, reynolds2023salient, tseng2022vizwiz,  zheng2023materobot} to showcase the feasibility of their concepts to assist visually impaired. Our work falls into the latter case.

\section{Method: Open Scene Understanding}
\label{sec:methodology}
\begin{figure*}
    \centering
    \includegraphics[width=0.99\textwidth]{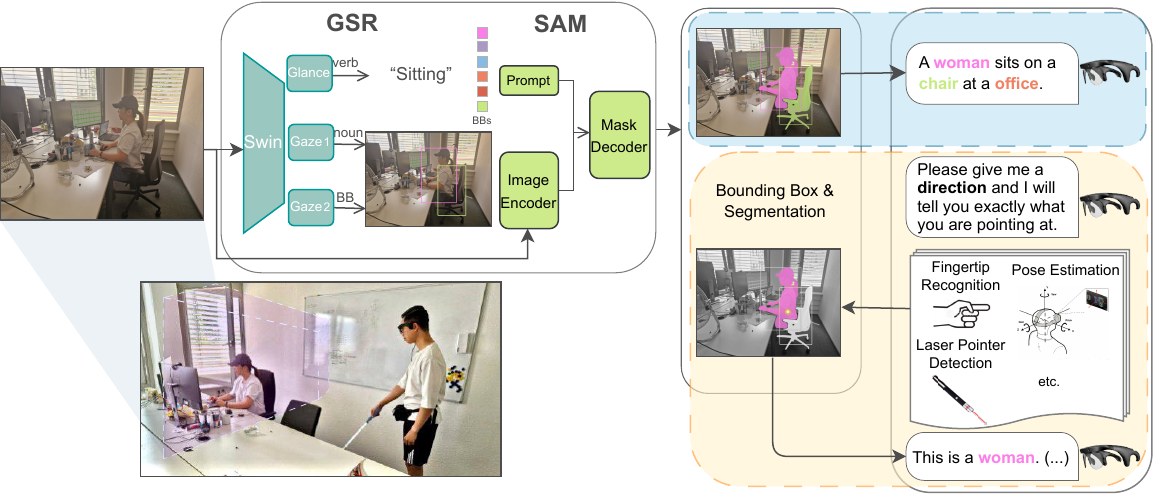}
    \caption{\textbf{The overview of Open Scene Understanding (OpenSU) system}. Grounded situation recognition captures the activity (\eg, \emph{sitting}), nouns (\eg, \textcolor{magenta}{woman}, \textcolor{LimeGreen}{chair}, \textcolor{red}{office}) related to the roles (Agent, Item, Place), and the bounding boxes of the objects. 
    The caption template is ``An \underline{Agent} sits on an \underline{Item} at a \underline{Place}'', so the image caption of the verb \emph{sitting} is ``A woman sits on a chair at an office''. SAM uses bounding boxes as prompts to generate the segmentation masks. According to application interaction, potential region indication methods (\eg, fingertip, head pose, laser pointer) can be used to specify a region of interest, and the information can be reported to the user via bone-conducting earphones of wearable systems~\cite{zhang2021trans4trans}.}
    \label{fig:pipeline}
\end{figure*}

As shown in Fig.~\ref{fig:pipeline}, our Open Scene Understanding (OpenSU) system consists of three parts: 
(1) the grounded situation recognition;
(2) the segmentation map generation; 
(3) the information transformation.
A rough image caption, including main activities and objects, and bounding boxes of the objects are produced by the grounded situation recognition model. Since the objects interact with each other, the overlapping bounding boxes can not classify the object in the region of interest properly. To address this, the Segment Anything Model (SAM) is adopted behind the GSR model to obtain the non-overlapping segmentation masks annotated with roles related to the activities and the object class. Finally, we propose several region indication methods to easily specify an object of interest by PVI.

\subsection{Grounded Situation Recognition}
\label{sec:3.1}
\noindent\textbf{Preliminary of CoFormer.}
Our GSR Model is based on CoFormer~\cite{cho2022coformer}, which is inspired by Kahneman's cognitive theory~\cite{kahneman2003cognitivesystem}. The theory suggests that humans employ two comprehensive thinking systems to make decisions. In the case of the scene understanding task, humans rapidly glance at a scene to perceive the overall situation, then concentrate on specific details to analyze the objects involved and their relations. CoFormer builds upon the DETR~\cite{carion2020detr} model, a significant milestone in end-to-end object detection, to simultaneously predict the activity, salient objects, and their corresponding bounding boxes. The image feature map extracted from CNN~\cite{he2016resnet} backbone is processed with three transformer decoders. The first decoder, the Glance Transformer extracts holistic information and predicts the activity (verb) depicted in the scene. Based on the predicted verb, the set of roles is determined. The second decoder, Gaze-S1 captures the nouns and their relations using learnable role tokens as input. The third decoder, Gaze-S2 takes the aggregated image features and learnable tokens, predicted verb, and its associated roles as input. It extracts role features that are crucial for predicting grounded nouns. 

\noindent\textbf{Swin Transformer for GSR.}
CoFormer performed state-of-the-art on the grounded scene understanding dataset, SWiG~\cite{pratt2020swig}. However, since the multi-stream transformer decoder makes predictions only based on the output feature map from the backbone, we argue that the capability of the backbone in extracting high-level image features greatly impacts the overall performance of CoFormer. According to Dosovitskiy \textit{et al.}~\cite{dosovitskiy2020image}, CNNs exhibit inferior performance in modeling global context due to the restricted receptive fields of the convolution operation, while vision transformers extract both global and local information through the self-attention mechanism, delivering the ability to capture long-range dependencies effectively. 
To include the long-range modeling capacity for GSR, we utilize the lightweight version of Swin Transformer~\cite{liu2021swin} pre-trained on ImageNet-1k~\cite{deng2009imagenet} to replace the CNN backbone.
This design is non-trivial, since 
Swin Transformer 
shows superiority to extract the local representation through local window attention and the interaction among regions through shifted window attention. Performing self-attention computations within multiple windows is significantly more computationally efficient compared to applying global self-attention. Current GSR models~\cite{cho2022coformer} apply multiple decoders to extract both global (activity) and local (entities) information from the feature map obtained by the CNN backbone. 
In contrast, Swin Transformer which produces feature maps containing both global and local information is better suited for the GSR task than CNN. More results are presented in Sec.~\ref{sec:4.3}.

\noindent\textbf{GELU for GSR.}
Additionally, we accelerate the convergence by replacing the ReLU~\cite{nair2010relu} activation function with GELU~\cite{hendrycks2016gelu}, a non-linear approximation of ReLU with the adaptive dropout feature, in the decoder transformers and the classifiers. Since ReLU can guarantee the gradient on the activation function to be constant $1$ or $0$ during backpropagation, it solves the gradient vanishing issue and reduces the huge amount of operations caused by updating the gradient with tiny values in practical engineering problems. It simplifies the complexity of computation and brings the effect of sparse activation, which is competitive in deep neural networks. ReLU~\cite{nair2010relu} is formulated as:
\begin{equation}
    \text{ReLU}(x) = \max\left(0,x\right).
\end{equation}
However, as a result of sparsity in ReLU, several neurons with ReLU activation function are in a DEAD state and the parameters cannot be updated. Instead, this leads to a decrease in the representation capability, resulting in a certain loss of performance as well as training speed. Moreover, the ReLU outputs remain unbounded on the positive side, leading to potential issues like gradient explosion during training. In contrast, GELU looks at the non-linearity problem from a probabilistic view and delivers a continuous differentiable output. It incorporates the benefits of both stochastic regularization and ReLU. On the basis of one-sided saturation, it maintains a certain negative correction capability by retaining certain values in the partially negative region.  GELU circumvents the issue of dead ReLU to a certain extent but retains the sparsity caused by dead ReLU. Additionally, the nonlinearity of GELU in the positive region enhances the fitting of complex nonlinear functions compared to ReLU, allowing the GSR model to capture more complex and nuanced relationships in the data. This leads to an advanced understanding of the grounded situation. To realize GELU, the input $x$ is multiplied with the standard Gaussian cumulative distribution function $\Phi\left(x\right)$:
\begin{equation}
\begin{aligned}
        \text{GELU}(x) &= xP\left(X\leq x\right)= x\Phi\left(x\right)\\
    &= x\cdot\frac{1}{2}\left[1+\text{erf}\left(x/\sqrt{2}\right)\right].
\end{aligned}
\end{equation}
GELU activation function is widely used in vision transformers~\cite{dosovitskiy2020image, liu2021swin, wang2021pvt}, but its performance is barely discussed. In our experiments (Sec.~\ref{sec:4.4}), we showcase the effectiveness of GELU in our pure transformer architecture for grounded situation recognition and scene understanding. 

\subsection{Segmentation Map Generation}\label{sec:3.2}
To gain the accurate boundaries of the salient objects, we leverage the original image as input and the bounding boxes outputted by the GSR as prompts for Segment Anything Models (SAMs).
SAM serves as a promptable foundation model for image segmentation, combining the benefits of both semantic and instance segmentation. It provides the capability to generate hundreds of masks per image without prompt or generate specific masks using sparse prompts such as points, boxes, or text, as well as dense prompts like masks. Typically, SAM utilizes bounding boxes as prompts for achieving instance segmentation, as masks generated by SAM with point prompts might represent minute components of an instance. 

The solid but heavyweight image encoder ViT-H~\cite{dosovitskiy2020image} with $611M$ parameters empowers Segment Anything to perform superior. Recent efforts~\cite{zhang2023mobilesam,zhao2023fastsam} dedicate to compress Segment Anything meanwhile remaining its robustness. MobileSAM~\cite{zhang2023mobilesam} shares the same architecture with Segment Anything and employs TinyViT~\cite{wu2022tinyvit} with $5M$ parameters as an image encoder. The knowledge extracted from SAM is transferred into MobileSAM via knowledge distillation. MobileSAM improves not only the inference efficiency but also the training efficiency. It is trained on a single GPU with $1\%$ of the original images for less than one day. Because of the limited computation resources of mobile assistance systems, it is crucial to perform efficient inference to enhance the user experience. Therefore, in this work, we spend a large effort to test and compare both vanilla SAM and MobileSAM in our OpenSU framework. 

\subsection{Region of Interest Indication}
Multiple sensors, like an RGB-D camera and Inertial Measurement Units (IMUs), are installed on smart glasses~\cite{zhang2021trans4trans}. They can be used to capture the indicator and infer the specified direction of interest.
Here, we introduce several potential methods to indicate the region of interest.
Please note that integrating different interaction methods in our system remains in our future work. 

Firstly, we can capture the gesture through cameras and use gesture recognition algorithms~\cite{alam2022unified, chen2021egocentric} to determine where the user is pointing. Gesture recognition is already involved in some VR/AR development kits, like Hololens. Furthermore, the work of Miksik \textit{et al.}~\cite{miksik2015laserpointer} on laser pointer detection coupled with optical see-through glasses (EPSON MOVERIO BT-200) serves as an inspiration for our direction indication. The utilization of IMUs enables the outside-in 6 DOF pose tracking and pose estimation in AR glasses, as analyzed by Firintepe~\textit{et al.}~\cite{firintepe2021poseestimation}.  The system output is focused on identifying the central object within the scene.

\section{Experiments}
\label{sec:experiments}
\subsection{Dataset and Metrics}
\noindent\textbf{Dataset.} 
The GSR dataset, SWiG~\cite{pratt2020swig} is extended from the situation recognition dataset imSitu~\cite{yatskar2016imsitu} by adding $278,336$ bounding boxes. Each sample of imSitu is labeled with a \textit{verb}, which is associated with up to $6$ \textit{roles}. Three workers annotate the \textit{entities} and their \textit{bounding boxes} for the images using Amazon’s Mechanical Turk framework. The \textit{bounding boxes} are described with format $\left[x1, y1, x2, y2\right]$.
SWiG dataset contains $75K$/$25K$/$25K$ images for training/validation/testing. The total numbers of \textit{verb}, \textit{roles} and \textit{entities} classes are $504$, $190$ and $11,538$. The activity-role pairs are derived from FrameNet~\cite{baker1998framenet} and the entity classes source from ImageNet~\cite{deng2009imagenet}.
\begin{table*}[t]
\centering
\small
\caption{\small \label{tab:swig_dev_set_result} \textbf{Results (\%) on the SWiG dataset}, including three settings and five metrics evaluated on the val and the test set.}
\vspace{-3mm}
\resizebox{1.99\columnwidth}{!}{
\begin{tabular}{l|ccccc|ccccc|cccc}
\hline
\multicolumn{1}{c|}{\multirow{2}{*}{Models}} & \multicolumn{5}{c|}{Top-1-Verb}      & \multicolumn{5}{c|}{Top-5-Verb}      & \multicolumn{4}{c}{Ground-Truth-Verb}              \\
\multicolumn{1}{c|}{} & verb & value& val-all        & grnd & grnd-all       & verb & value& val-all        & grnd & grnd-all       & value& val-all        & grnd  & grnd-all \\ \hline
\hline
\rowcolor{gray!15} \multicolumn{15}{c}{Methods for Grounded Situation Recognition ({val} set)} \\  \hline  
ISL~\cite{pratt2020grounded}        & 38.83& 30.47& 18.23& 22.47& 7.64 & 65.74& 50.29& 28.59& 36.90& 11.66& 72.77& 37.49& 52.92 & 15.00    \\
JSL~\cite{pratt2020grounded}         & 39.60& 31.18& 18.85& 25.03& 10.16& 67.71& 52.06& 29.73& 41.25& 15.07& 73.53& 38.32& 57.50 & 19.29    \\
GSRTR~\cite{cho2021grounded}       & 41.06& 32.52& 19.63& 26.04& 10.44& 69.46& 53.69& 30.66& 42.61& 15.98& 74.27& 39.24& 58.33 & 20.19    \\
SituFormer~\cite{wei2021rethinking} & 44.32 & 35.35 & 22.10 & 29.17 & 13.33 & 71.01 & 55.85 & 33.38 & 45.78 & 19.77 & 76.08 & 42.15 & 61.82 & 24.65    \\
CoFormer~\cite{cho2022coformer} & 44.41 & 35.87 & 22.47 & 29.37 & 12.94 & 72.98 & 57.58 & 34.09 & 46.70 & 19.06 & 76.17 & 42.11 & 61.15 & 23.09\\
GSRFormer~\cite{cheng2022gsrformer} & {46.64} & {37.69} & {23.58} & {31.61} & {14.42} & {73.43} & {58.75} & {35.82} & {48.42} & {21.67} & \bf{78.76} & {44.71} & {63.95} & {25.85}\\
\bf{OpenSU} & \bf{49.96} & \bf{41.12} & \bf{26.69} & \bf{34.69} & \bf{16.27} & \bf{77.77} & \bf{62.81} & \bf{39.23} & \bf{52.56}& \bf{23.50}& {78.68} & \bf{46.66}& \bf{65.33} & \bf{27.66}\\ \hline
\rowcolor{gray!15} \multicolumn{15}{c}{Methods for Grounded Situation Recognition ({test} set)} \\  \hline 
ISL~\cite{pratt2020grounded}     & 39.36 & 30.09 & 18.62   & 22.73 & 7.72     & 65.51 & 50.16 & 28.47   & 36.60 & 11.56    & 72.42  & 37.10    & 52.19  & 14.58    \\
JSL~\cite{pratt2020grounded}     & 39.94 & 31.44 & 18.87   & 24.86 & 9.66     & 67.60 & 51.88 & 29.39   & 40.60 & 14.72    & 73.21  & 37.82    & 56.57  & 18.45    \\
GSRTR~\cite{cho2021grounded}    & 40.63 & 32.15 & 19.28   & 25.49 & 10.10    & 69.81 & 54.13 & 31.01   & 42.50 & 15.88    & 74.11  & 39.00    & 57.45  & 19.67    \\
SituFormer~\cite{wei2021rethinking}& 44.20 & 35.24 & 21.86   & 29.22 & 13.41    & 71.21 & 55.75 & 33.27   & 46.00 & 20.10    & 75.85  & 42.13    & 61.89  & 24.89    \\
CoFormer~\cite{cho2022coformer} & 44.66 & 35.98 & 22.22 & 29.05 & 12.21 & 73.31 & 57.76 & 33.98 & 46.25 & 18.37 & 75.95 & 41.87 & 60.11 & 22.12\\
GSRFormer~\cite{cheng2022gsrformer} & {46.53} & {37.48} & {23.32} & {31.53} & {14.23} & {73.44} & {58.84} & {35.82} & {48.43} & {21.41} & \bf{78.81} & {44.68} & {63.87} & {25.35}\\
\bf{OpenSU} & \bf{50.10}&\bf{41.20}&\bf{26.56}&\bf{34.27}&\bf{15.70}&\bf{77.91}&\bf{62.95}&\bf{39.00}&\bf{51.90}&\bf{22.74}&{78.67}&\bf{46.31}&\bf{64.22}&\bf{26.57}\\
\hline
\end{tabular}}
\vspace{-3mm}
\end{table*}

\begin{table}
\centering
\footnotesize
\caption{\textbf{Efficiency analysis} of scene understanding methods and Segment Anything Model variants.}
\vskip -2ex
\label{tab:run_time}
\resizebox{\columnwidth}{!}{
\setlength{\tabcolsep}{1mm}
\begin{tabular}{l@{}cc} 
\toprule
\textbf{System}       & \textbf{Runtime(s)} & \textbf{\#Params(M)}\\\midrule\midrule
Grounding DINO~\cite{liu2023GroundedDINO}+SAM+BLIP~\cite{li2023blip} & 74.04& 1310.66\\
OpenSU (SAM) &1.34& 738.09\\
OpenSU (MobileSAM) & \textbf{0.70}&\textbf{107.13}\\

\bottomrule
\end{tabular}
}
\vskip -2ex
\end{table}
\noindent\textbf{Metrics.} 
Five metrics are reported to evaluate the GSR models' performance: (1) \textbf{verb} to measure verb prediction accuracy, (2) \textbf{value} to assess the accuracy of a predicted \textit{noun} for a given \textit{role}, (3) \textbf{value-all} to measure the correctness of all \textit{nouns} within a frame,  (4) \textbf{grounded-value} to determine if the \textit{noun} is accurately predicted and its bounding box has an IoU over $0.5$, and (5) \textbf{grounded-value-all} to measure the frequency that all the \textit{noun-bounding box} pairs are correct. Because of the structured annotation, all metrics are highly related to the \textit{verb} accuracy. And several activities could take place in a scene. Three settings are proposed to evaluate the predictions comprehensively. (1) \textbf{Top-1 Verb}: once the top-1 predicted \textit{verb} is not the ground truth \textit{verb} of the sample, all the \textit{verb}, \textit{noun}, and \textit{bounding boxes} are regarded as wrong. (2) \textbf{Top-5 Verb}: the \textit{verb} is correct, when the ground truth \textit{verb} is contained in the \textit{verbs} with top-5 probabilities. (3) \textbf{Ground-Truth Verb}: the ground truth \textit{verb} is assumed to be given in this case.

\subsection{Implementation Details}
Our model is trained with a batch size of $4$ for $40$ epochs on four Tesla V100-SXM2-32GB GPUs. Each input image is padded to size $700\times700$. The transformer backbone Swin-Tiny~\cite{liu2021swin} is pre-trained on ImageNet-1k~\cite{deng2009imagenet}. We adopt AdamW optimizer with $10^{-4}$ weight decay, $\beta_1 {=} 0.9$ and $\beta_2 {=} 0.999$. 
The learning rates are different for backbone and transformer decoders, \ie, $10^{-5}$ and $10^{-4}$. After epoch $30$, the learning rates decrease with factor $10$. We use the same loss functions as CoFormer~\cite{cho2022coformer} to align logits and targets, which are verb classification loss, noun classification loss, box existence loss, and box regression loss.
To compare the inference time, 10 images with size $1042\times1042$ go through our OpenSUs and other frameworks. The average runtime per image and the model parameters are calculated. The evaluation process is conducted on one Tesla V100-SXM2-32GB GPU.
\begin{figure}[!t]
    \centering
    \includegraphics[width=0.47\textwidth]{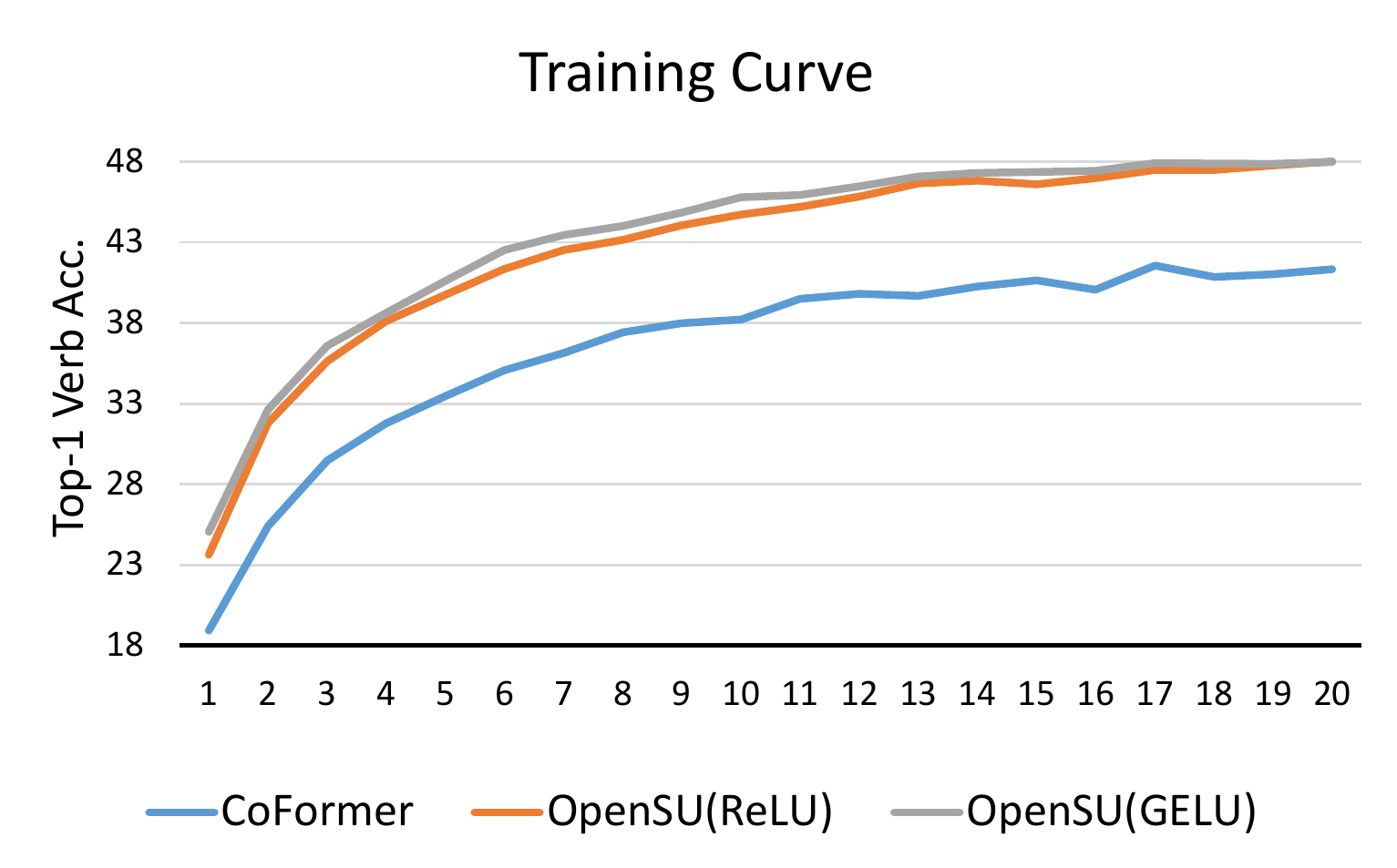}
    \vskip -2ex
    \caption{\textbf{Training curves} of the GSR model, CoFormer~\cite{cho2022coformer}, and our OpenSU systems, considering both ReLU and GELU as the decoder activation functions.}
    \label{fig:training_curve}
    \vskip -2ex
\end{figure}

\subsection{Results on SWiG}\label{sec:4.3}
In this experiment, we focus on comparing the performance of our proposed method with other state-of-the-art methods. The quantitative results are illustrated in Tab.~\ref{tab:swig_dev_set_result}, which indicates the proposed OpenSU outperforms other counterparts among all evaluation metrics except for the value metric in the Ground-Truth Verb setting. Compared with the second best method~\cite{cheng2022gsrformer}, OpenSU achieves a maximum $4.34\%$ performance gain with the verb metric in Top-5-Verb setting on the validation set, and minimum $1.39\%$ boost with the grounded-value metric in Ground-Truth-Verb setting. Similar improvements can be found on the SWiG test set with maximum $4.47\%$ and minimum $0.36\%$ referring to the verb metric in Top-5-Verb and the grounded-value metric in the Ground-Truth-Verb setting, respectively. As for the value metric in the Ground-Truth-Verb setting, OpenSU achieves very close performance to GSRFormer~\cite{cheng2022gsrformer}, the most competitive counterpart listed in Tab.~\ref{tab:swig_dev_set_result}. It is worth noting that the performance gap between the validation and test set is very small, which showcases our OpenSU has an excellent generalization ability.

\begin{figure*}
    \centering
    \includegraphics[width=0.99\textwidth]{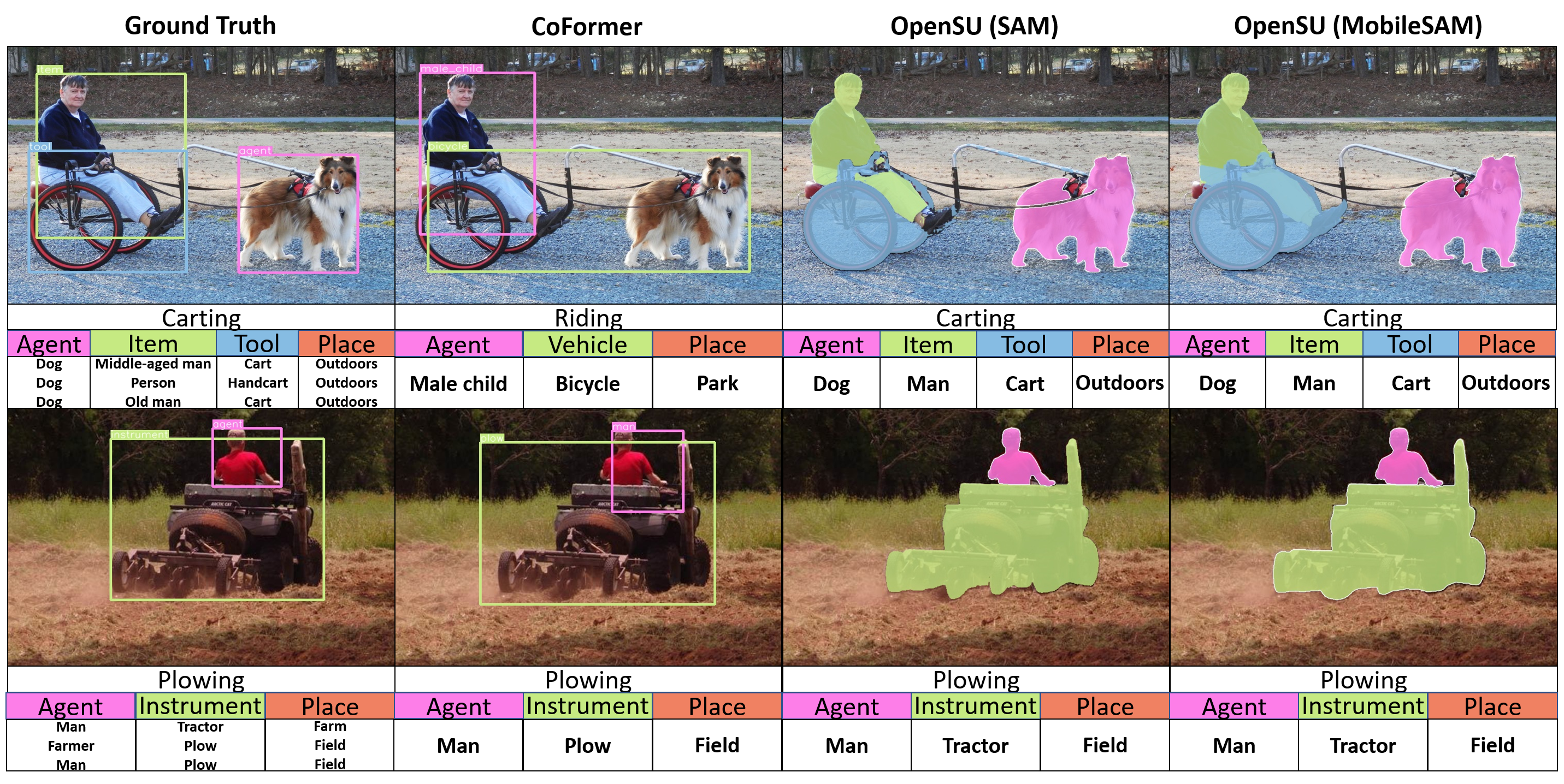}
    \vskip -2ex
    \caption{\textbf{Qualitative comparison} between GSR ground truth, predictions obtained from the GSR baseline CoFormer~\cite{cho2022coformer}, and fine-grained predictions of two SAM-based variants of our OpenSU system.}
    \label{fig:qualitative}
    \vskip -2ex
\end{figure*}

\subsection{Efficiency Analysis}\label{sec:4.4}

We further conduct experiments to analyze the efficiency. 
Tab.~\ref{tab:run_time} showcases the runtime in seconds and the number of parameters of all systems. It can be easily found that the MobileSAM-based OpenSU has the lowest number of parameters, which is less than one-tenth ($-1203.53$ MParams) compared with the combination of Grounding DINO~\cite{liu2023GroundedDINO}, SAM and BLIP~\cite{li2023blip}. Besides, OpenSU with MobileSAM can achieve the smallest runtime, \ie, $0.70s$, reducing $73.34s$ compared to the combination baseline. Furthermore, Fig.~\ref{fig:training_curve} presents the convergence speed of models using different activation functions. Our OpenSU with GELU converges faster than OpenSU with RELU and CoFormer~\cite{cho2022coformer} while keeping the highest verb accuracy under the Top-1-Verb setting during training. This evidence proves that the smoothness and continuity of GELU facilitate more stable and efficient optimization during training.

\subsection{Qualitative Results}
Qualitative results of CoFormer~\cite{cho2022coformer} and our OpenSUs are presented in Fig.~\ref{fig:qualitative}. For each image, there are three frames that originate from three annotators, each of which can be regarded as ground truth.
In the first example, since CoFormer wrongly predicts the activity, the related information including roles, entity classes, and bounding boxes is incorrect as well. Our OpenSU systems with the state-of-the-art GSR model obtain the correct text and spatial information. Compared to OpenSU with MobileSAM, where the upper body and the legs are delineated as separate entities, OpenSU with original SAM can segment the details of the overlapping objects. 
The second row 
showcases another situation. All three models predict the correct verb and nouns. 
However, the positions of the bounding boxes predicted by CoFormer have a serious shift and the large area of overlap leads to confusion. Both variations of the proposed OpenSU output the corresponding segmentation masks instead of the bounding boxes resulting in more precise prediction and better understanding.

\subsection{Field Test}
\begin{figure}
    \centering
    \includegraphics[width=0.5\textwidth]{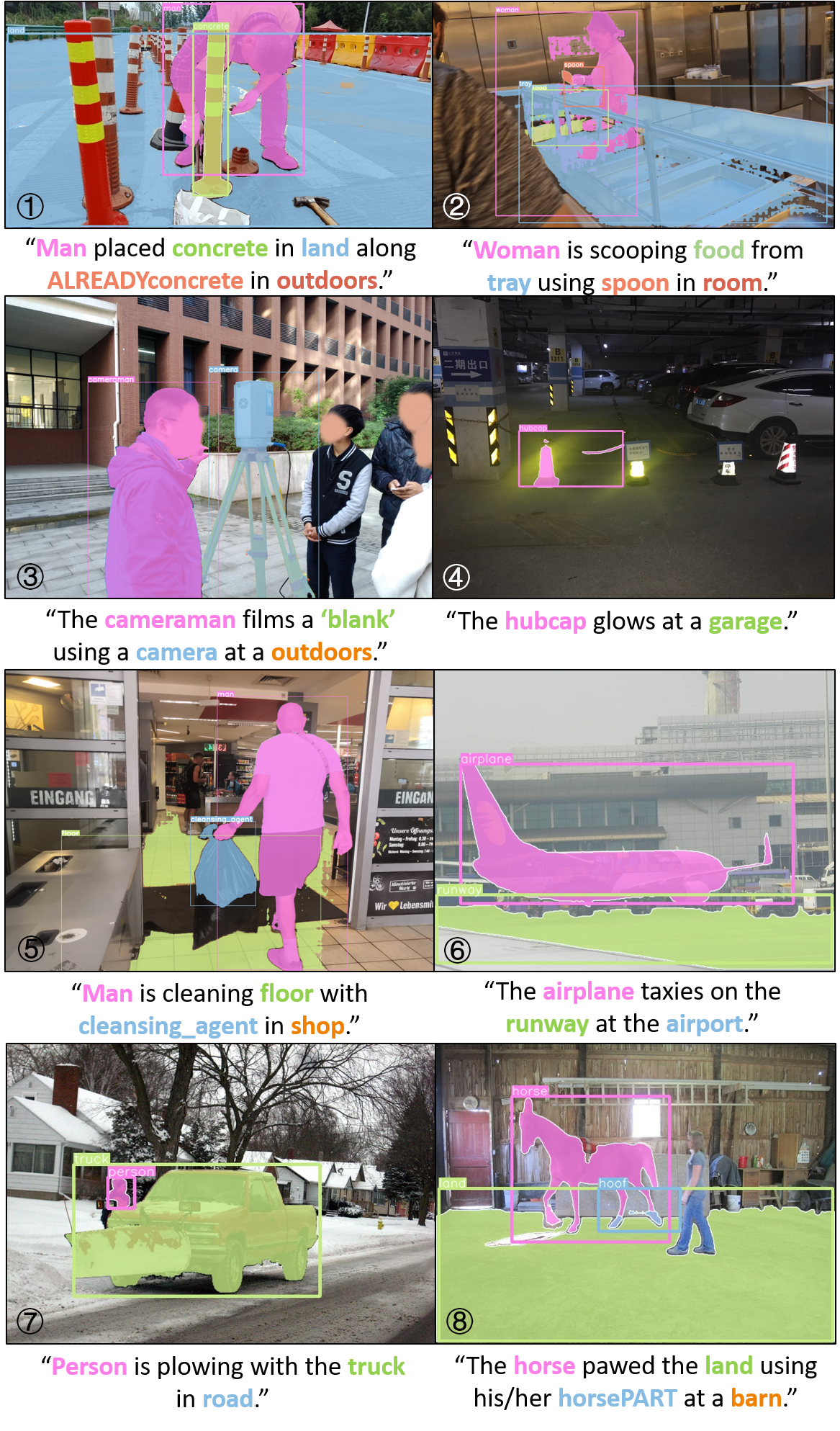}
    \vskip -2ex
    \caption{\textbf{Field test with the OpenSU system}. The samples are derived from datasets specifically proposed for assisting the visually impaired, \ie, OD~\cite{tang2023do} and WOTR~\cite{XIA20231wotr}.}
    \label{fig:field_test}
    \vskip -3ex
\end{figure}

Apart from the evaluation on the SWiG dataset, the OpenSU system underwent a field test on two visually impaired assistance datasets, \ie, Obstacle Dataset (OD)~\cite{tang2023do} and Walk On The Road (WOTR)~\cite{XIA20231wotr}. The field test encompassed comprehensive daily scenes, allowing for a thorough evaluation of the system performance in assisting PVI. Fig.~\ref{fig:field_test} visualizes the results of OpenSU, including \normalsize{\textcircled{\scriptsize{1}}}\normalsize~construction site, \normalsize{\textcircled{\scriptsize{2}}}\normalsize~dining hall, \normalsize{\textcircled{\scriptsize{3}}}\normalsize~campus, \normalsize{\textcircled{\scriptsize{4}}}\normalsize~garage, \normalsize{\textcircled{\scriptsize{5}}}\normalsize~supermarket, \normalsize{\textcircled{\scriptsize{6}}}\normalsize~airport, \normalsize{\textcircled{\scriptsize{7}}}\normalsize~road,
and \normalsize{\textcircled{\scriptsize{8}}}\normalsize~farm.
Accurate segmentation of entities can be achieved even when the samples (\normalsize{\textcircled{\scriptsize{1}}}\normalsize~and \normalsize{\textcircled{\scriptsize{2}}}\normalsize) contain five interactive roles. The detection of the woman and garbage in~\normalsize{\textcircled{\scriptsize{2}}}\normalsize~and~\normalsize{\textcircled{\scriptsize{5}}}\normalsize~is affected by the objects with the same class and same color respectively, leading to bounding boxes with incorrectly predicted sizes. Overlapping bounding boxes do not degrade segmentation performance, but inaccurate positioning and sizing can reduce accuracy.
The absence of human agents presented in the samples \normalsize{\textcircled{\scriptsize{4}}}\normalsize, \normalsize{\textcircled{\scriptsize{6}}}\normalsize, and \normalsize{\textcircled{\scriptsize{8}}}\normalsize~signifies that GSR differs from Human-Object Interaction (HOI) and action recognition tasks. 
These tasks are generalized by introducing verbs, such as ``glowing'', ``taxiing'', and ``pawing'', to specifically describe objects' activities instead of people's actions.
We observe that the descriptions of roadblocks are incorrect in \normalsize{\textcircled{\scriptsize{1}}}\normalsize~and \normalsize{\textcircled{\scriptsize{4}}}\normalsize. As part of future work, we suggest involving additional information about roadblocks in the dataset to improve accuracy. 
Based on the field test, the image captions provided in our analysis are considered rough, as they consist of predefined frames for each verb with entities filled in. However, despite their roughness, people can comprehend them due to the structured image description.

\subsection{Discussion}
\noindent\textbf{(1) What can AGI promote assistive technology?}
Artificial General Intelligence (AGI) has the potential to promote the development of assistive technology, 
including scene perception and understanding, scene reasoning, adaptive learning for specific needs and preferences of users, high-level decision-making in mobility and navigation, etc. These capabilities can be leveraged to develop more intelligent and adaptive assistive technologies, enabling them to better respond to the special needs of users, particularly for people with disabilities. However, the recent AGI technology is resource-intensive. Thus, making AGI-based assistance systems accessible and affordable to a wide range of users, poses a significant challenge. Therefore, improving efficiency is crucial to enhance the overall user experience. 

\noindent\textbf{(2) How to leverage big vision models to help PVI?}
By leveraging the advantages of big vision models, \ie, Segment Anything Model~(SAM), we enhance the object recognition and positioning of GSR by delivering fine-grained segmentation masks instead of ambiguous bounding boxes. This approach enables more precise interaction experiences when navigating and interacting with known or unknown environments. The proposed OpenSU system is the first attempt to combine SAM with GSR for helping PVI. Moving forward, the integration of SAM in various assistive applications will be further explored to unlock its full potential, such as customized object recognition, prompt-based object finding, path finding, social distancing, \textit{etc.}

\section{Conclusion}
\label{sec:conclusion}
People with Visual Impairments (PVI) often encounter challenges in understanding scenes in their daily lives. In this work, we 
propose an open scene understanding system (OpenSU) to assist PVI 
to obtain more comprehensive scene descriptions. 
Specifically, we utilize the Swin transformer as the backbone, enabling more accurate Grounded Situation Recognition (GSR). Besides, we for the first time, adopt the Segment Anything Model (SAM) to generate finer-grained masks for more precise object positioning compared to traditional bounding boxes of GSR. Through field testing, OpenSU showcases the effectiveness and the advantage of using SAM and GSR to perform open scene understanding in daily scenarios.

\clearpage
{\small
\bibliographystyle{ieee_fullname}
\bibliography{egbib}
}

\end{document}